\pdfoutput=1

\documentclass[11pt]{article}

\usepackage{acl}

\usepackage{times}
\usepackage{latexsym}

\usepackage[T1]{fontenc}

\usepackage[utf8]{inputenc}

\usepackage{microtype}

\usepackage{inconsolata}

\usepackage{graphicx}

\usepackage{hyperref}
\usepackage{url}
\usepackage{booktabs}

\usepackage{lineno}

\definecolor{darkblue}{rgb}{0, 0, 0.5}
\hypersetup{colorlinks=true, citecolor=darkblue, linkcolor=darkblue, urlcolor=darkblue}
\usepackage{float}

\usepackage{booktabs}
\usepackage[most]{tcolorbox}
\usepackage{soul}
\usepackage{multirow}
\usepackage{xspace}
\usepackage{subcaption}
\usepackage{enumitem}
\usepackage{cleveref}
\usepackage{tabularx}
\crefname{section}{§}{§§}
\Crefname{section}{§}{§§}
\usepackage{comment}

\makeatletter
\newcommand\thefontsize[1]{{#1 The current font size is: \f@size pt\par}}
\makeatother

\newcommand{\papercomment}[3]{{\textcolor{#3}{[#1 #2]}}}
\newcommand{\marker}[1]{\textbf{#1:} }
\newif\ifcomments
\ifcomments
    \newcommand{\rob}[1]{\papercomment{\marker{rob}}{#1}{cyan}}
    \newcommand{\sameer}[1]{\papercomment{\marker{sameer}}{#1}{green!50!black}}
    \newcommand{\joel}[1]{\papercomment{\marker{joel}}{#1}{purple}}
    \newcommand{\ganesh}[1]{\papercomment{\marker{ganesh}}{#1}{violet}}
    \newcommand{\tamanna}[1]{\papercomment{\marker{tamanna}}{#1}{magenta}}
\else
    \newcommand{\rob}[1]{}%
    \newcommand{\sameer}[1]{}%
    \newcommand{\joel}[1]{}%
        \newcommand{\ganesh}[1]{}%
    \newcommand{\tamanna}[1]{}%
\fi
\newcommand{\sep}{\hspace{0.75em}}

\title{Characterizing Mamba's Selective Memory using Auto-Encoders}

\author{
 \textbf{Tamanna Hossain\thanks{Corresponding author. Work done while an intern at Dataminr Inc.}\textsuperscript{\textdagger}}
 \sep
 \textbf{Robert L. Logan IV\textsuperscript{\textdaggerdbl}}
 \sep
 \textbf{Ganesh Jagadeesan\textsuperscript{\textdaggerdbl}}
\\
 \textbf{Sameer Singh\textsuperscript{\textdagger}}
 \sep
 \textbf{Joel Tetreault\textsuperscript{\textdaggerdbl}}
 \sep
 \textbf{Alejandro Jaimes\textsuperscript{\textdaggerdbl}}
\\
 \textsuperscript{\textdagger}University of California, Irvine
 \sep
 \textsuperscript{\textdaggerdbl}Dataminr Inc.
\\
   \texttt{\{tthossai,sameer\}@uci.edu}
   \sep
   \texttt{aj27@caa.columbia.edu}
 \\
   \texttt{\{rlogan,cjagadeesan,jtetreault\}@dataminr.com}
}

\begin{document}

\maketitle
\begin{abstract}

State space models (SSMs) are a promising alternative to transformers for language modeling because they use fixed
memory during inference.
However, this fixed memory usage requires some information loss in the hidden state when processing long sequences.
While prior work has studied the sequence length at which this information loss occurs, it does not characterize the types of information SSM language models (LMs) tend to forget.
In this paper, we address this knowledge gap by identifying the types of tokens (e.g., parts of speech, named entities) and sequences (e.g., code, math problems) that are more frequently forgotten by SSM LMs.
We achieve this by training an auto-encoder to reconstruct sequences from the SSM's hidden state, and measure information loss by comparing inputs with their reconstructions.
We perform experiments using the Mamba family of SSM LMs (130M--1.4B) on sequences ranging from 4--256 tokens.
Our results show significantly higher rates of information loss on math-related tokens (e.g., numbers, variables), mentions of organization entities, and alternative dialects to Standard American English.
We then examine the frequency that these tokens appear in Mamba's pretraining data and find that less prevalent tokens tend to be the ones Mamba is most likely to forget.
By identifying these patterns, our work provides clear direction for future research to develop methods that better control Mamba's ability to retain important information.%
\footnote{Our code is available at: \url{https://github.com/dataminr-ai/ouroboros}.}

\end{abstract}

\section{Introduction}

\begin{figure}[!th]
    \centering
    \includegraphics[width=\linewidth]{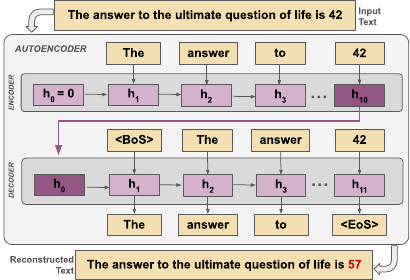}
    \caption{\textbf{Identifying Selective Memory Loss.} We train auto-encoders to reconstruct inputs directly from Mamba’s hidden states, and measure information loss by comparing inputs to their reconstructions.
    These reconstructions act as probes of the hidden state's information retention: more faithful reconstruction implies greater information retention.
    }
    \label{fig:recon}
    
\end{figure}

State space models (SSMs) have emerged as a promising alternative to transformers~\citep{Vaswani2017AttentionIA} for language modeling, with the Mamba~\citep{Gu2023MambaLS} SSM LMs achieving comparable performance to similar sized transformer models while being substantially more memory efficient at inference time.
This memory efficiency arises from the fact that SSM next token predictions are computed recurrently using a fixed-size hidden state, whereas transformer next token predictions are computed using a key-value cache that grows linearly with the sequence length.

Recent work has shown that this efficiency comes at the cost of some information loss when processing long sequences.
In particular, \citet{Jelassi2024RepeatAM} establish an
upper bound on the length of sequences that can be exactly copied by Mamba, while \citet{wang2025understanding} prove that the influence of an input token decays exponentially with input length.
To provide empirical support for these theoretical results, these works measure Mamba's decline in accuracy as sequence length increases on copying tasks that either use synthetic inputs, or prime the model to memorize information using cues in the prompt.
While these results characterize \emph{when} Mamba begins to forget, they leave open the question of \emph{what} kinds of information is more likely to be forgotten, particularly on natural inputs.

In this paper, we develop a methodology that allows us to better understand the types of tokens and sequences that the Mamba LM is likely to forget in practical settings.
The core idea of our approach is to compare input texts with known properties---such as named entity recognition (NER) and parts of speech (POS) tags on the token level, and categorical labels on the sequence level---to reconstructions obtained from Mamba's hidden state.
By correlating reconstruction errors (measured using F1 scores and token omission rates) with the input properties we are able to obtain high-level insights about Mamba's \emph{selective memory}.
For example, in Figure~\ref{fig:recon} we see that the number "42" is reconstructed as "57", providing some evidence that Mamba has a tendency to forget numbers.

To obtain input reconstructions, we train an auto-encoder that decodes sequences from hidden states produced by a frozen Mamba encoder.
By using pretrained checkpoints for our encoder, we are able to examine the impact language model pretraining on Mamba's memory, rather than just the architecture.
This approach
additionally complements the approaches used by prior works in that it allows us to study reconstructions of \emph{any} piece of text without having to provide memorization cues.
In Section~\ref{sec:validation} we measure the performance of this approach for a number of different model sizes and sequence lengths, and validate that it replicates findings from previous works.

In Sections~\ref{sec:token-level} and~\ref{sec:sequence-level} we then employ this approach to study token- and sequence-level characteristics of reconstruction errors on a number of datasets.
We begin by looking at Mamba's pretraining dataset The Pile~\citep{pile}, which allows us to understand how in-distribution errors vary by the input source.
Our results show that Mamba is significantly more likely to forget tokens on math-related sequences, even for relatively short sequence lengths (16+ tokens).
We also perform experiments on \citet{groenwold-etal-2020-investigating}'s SAE/AAVE tweet pair dataset, and find significantly elevated reconstruction errors on texts written using African American Vernacular English.

We then study token-level NER and POS errors on CoNLL-2003~\citep{tjong-kim-sang-de-meulder-2003-introduction}, and find significantly higher reconstruction errors on mentions of organizations, as well as numerical tokens.
To better understand this, we examine a 178M token sample from the Pile and find an association between the categories that Mamba tends to forget and the frequency that unique tokens belonging to those categories appear in the Pile, suggesting that their rarity in Mamba's pre-training data may contribute to their omission.
This raises the question of whether conventional pretraining pipelines are well suited for the inductive biases of SSM-based language models.
In sum, our work establishes clear differences in the types of information the Mamba LM tends to retain versus  forget, providing direction for future research into methods to improve the retention of important information in its hidden state.

\section{Related Work}

\paragraph{State Space Models} 
State space models (SSMs) are linear recurrent architectures with fixed-size hidden states that enable linear-time training and constant-time inference per step, making them a more efficient alternative to transformers for sequence modeling \citep{gu2021combining, gu2022efficiently}.
Recent models like Mamba \citep{Gu2023MambaLS} have shown that SSMs can match transformers of similar scale on standard NLP benchmarks, aided by long-range initialization techniques from HiPPO theory \citep{Gu2022HowTT} and time-dependent parameterization that allows selective attention over inputs.
Hybrid transformer-SSM architectures have shown gains in both efficiency and accuracy over pure transformers \citep{Wang2025M1TS}, reinforcing the value of SSM-based models—especially as inference-time scaling is increasingly used to enhance LM reasoning through techniques like task decomposition \citep{Wei2022ChainOT, Yao2023TreeOT} and increased sampling \citep{Wang2022SelfConsistencyIC}.

Although Mamba models offer improved computational efficiency over transformers, recent studies have shown that this comes at the cost of information degradation over long contexts. 
\citet{Jelassi2024RepeatAM} derive a theoretic upper bound on the sequence length that Mamba can recall information from, and \citet{wang2025understanding} demonstrate that token influence decays exponentially with input distance, i.e., there is a \emph{recency bias} to Mamba's memory. 
These findings are empirically supported by experiments using input sequences with explicit memorization cues and synthetic inputs \citep{Waleffe2024AnES, wang2025understanding}.
However, while these prior works illuminate \emph{when} forgetting occurs in Mamba's hidden state, it still remains unknown \emph{what} is forgotten, especially for natural inputs.
This is the knowledge gap we address in our paper.

\paragraph{Interpreting Hidden States} 
Our approach relates to a large body of work using learned classifiers to probe LM hidden states for the presence of properties such as syntactic and semantic knowledge (see \citet{rogers-etal-2020-primer} for an overview).
Our use of an auto-encoder to probe memorization of input sequences is unique in this literature, however this likely stems from the fact that input compression is not a concern for the transformer-based LMs typically considered in these works.
While auto-encoders have a rich history of use to train model hidden states to capture linguistic structures 
e.g., syntax and semantics  \citep{Zhang2023LearningDS}, negation and uncertainty \citep{Vasilakes2022LearningDR}, content and style \citep{li-etal-2022-variational-autoencoder, John2018DisentangledRL}, and tense, verb style and gender \citep{mercatali-freitas-2021-disentangling-generative}, there is little existing literature that uses auto-encoders to study hidden states for existing language models.
To our knowledge, the closest comparison to our work is \citet{templeton2024scaling}, who train sparse auto-encoders to learn discrete interpretable features captured by Claude's hidden states,\footnote{https://www.anthropic.com/claude}
however our work substantially differs from theirs in terms of both method and focus.

\section{Method}

In this section, we introduce a method for identifying the types of tokens and sequences Mamba is most likely to forget.
In our approach, we reconstruct text from Mamba’s hidden states using a trained auto-encoder and then compare the output to the original input annotated with features such as part-of-speech tags, named entities, and sequence-level categories.
This allows us to assess information retention in the natural usage of Mamba as a language model without relying on memorization prompts like prior works.

\subsection{Auto-Encoder}
Given an input text, $x$, the encoder function $f_{enc}(\cdot)$ maps the input into a latent space representation $\beta$.
 The decoder function $f_{dec}(\cdot)$ then takes this latent representation, $\beta$, and maps it back to the input space of text as $\tilde{x}$.
 Formally:

 $$\beta = f_{enc}(x),\,\tilde{x} = f_{dec}(\beta)$$

\paragraph{Encoder}
We use pretrained Mamba \citep{Gu2023MambaLS} language models with frozen weights%
\footnote{We use frozen weights in the encoder because our goal is to study the pre-trained Mamba checkpoints directly, evaluating the information preserved in their hidden states through standard language modeling.
}
to encode the input text into a latent space representation with two parts: (i) an \textit{SSM State}, $\beta_S$, and (ii) a \textit{Convolutional State}, $\beta_C$.
 
The Mamba architecture passes embeddings of an input sequence $x=[x_1,\ldots,x_n]$ through a convolutional filter, creating a convolutional hidden state, $\beta_C$.
After applying a non-linear activation we get an intermediate state, $z=[z_1,\ldots,z_n]$, which is passed into an SSM.
Within the SSM, recurrent states are computed as follows for time-dependent parameters $\bar{A}_t$ and $\bar{B}_t$ for $t \in \{1,\ldots,n\}$,

$$h_t = \bar{A}_th_{t-1} + \bar{B}_tz_t$$

The last recurrent state is taken as the SSM hidden representation of the input sequence, $\beta_S$.
Thus, the encoder maps an input sequence, $x$, into a latent state $\beta = [\beta_C , \beta_S]$.
 
\paragraph{Decoder} Once the latent representation $\beta$ is obtained from the frozen Mamba encoder, the decoder produces a reconstruction $\tilde{x}$ of the input text $x$.
In this work, we initialize the decoder using the same-pretrained Mamba architecture as the encoder.
The decoder (i) sets the initial state, $h_0$, of the decoder's SSM to the SSM state from the encoder, $\beta_S$, and (ii) similarly initializes the the decoder’s convolution with the convolutional state of the encoder, $\beta_C$. 
Given these initializations, the decoder autoregressively reconstructs the input text token by token.

\subsection{Measuring Information Loss} 
To measure information loss in Mamba’s hidden state, we compare input texts with their reconstructions produced by a trained auto-encoder. 
These reconstructions serve as probes into the information retention capacity of the hidden state: the more faithfully the input can be recovered, the more information must have been retained. 
We assess the fidelity of reconstructions using two metrics: (i) Omission Rate, and (ii) Rouge F1-Score.

{\bf Omission Rate} is a token-level metric that measures the frequency that specific input tokens are forgotten or omitted in reconstructions. 
Let $ f_{\text{in}}(t)$ and $f_{\text{rec}}(t)$ denote the frequency of token $t$ in the original input and reconstructed output, respectively. The omission rate for token $t$ is:

$$
\text{Omission Rate}(t) = 1 - \frac{f_{\text{rec}}(t)}{f_{\text{in}}(t)}
$$

An omission rate of 0 indicates that the perfect retention of a token, while a rate of 1 implies complete information loss. 

{\bf ROUGE F1-Score} \citep{lin-2004-rouge} provides a measure of reconstruction quality at the sequence level by measuring the overlap between the reconstructed text and the original input. 
Specifically, it balances \emph{precision}---how many of the reconstructed tokens are correct---with \emph{recall}---how many of the original tokens are reproduced. 

Together, these two metrics allow us to measure information loss at the token level (e.g., named entities, parts of speech) and the sequence level (e.g., math problems, emails). 
By measuring this information loss on subsets of data with known token- and sequence-level metadata (e.g., tokens with certain NER tags, or sequences from a particular source) we are able to extract high-level insights about the kinds of information Mamba is more likely to forget.

\section{Experiments and Results}

\subsection{Training}
We train the auto-encoder with Mamba models ranging from 130M-1.4B parameters to reconstruct sequences of lengths $l \in \{4,8,16,32,64,128,256\}$.
We initialize the autoencoders with pretrained Mamba checkpoints and freeze the encoder to ensure that we are evaluating the representations produced by the original model.

We train on the Pile---Mamba's pretraining dataset---to mitigate issues of distribution shift.
However, since the full Pile is no longer publicly available due to copyright restrictions, we use a version with all copyright-protected content removed, Pile Uncopyrighted.%
\footnote{\url{https://huggingface.co/datasets/monology/pile-uncopyrighted}}
We train auto-encoders to reconstruct texts at various fixed lengths in order to study how information is lost as the sequence length changes. 
To control for sequence length as a potential confounding factor in auto-encoder training and evaluation, we train separate auto-encoders for each fixed sequence length and evaluate them only on sequences of the corresponding length.

The encoder weights are frozen because the goal of our work is to study how Mamba, off the shelf, compresses information into its hidden state for its pre-training objective of next token prediction, while the decoder weights are trained specifically for the task of input reconstruction from the encoder’s hidden states.
The state sizes of different models are shown in Table \ref{table:size}.
We adopt a constant learning rate of 
$1\times10^{-5}$, matching the final learning rate of Mamba pre-training. 
We train until convergence with a batch size calibrated to the model size and GPU memory constraints.
We use 200K instances (282M tokens) from Pile Uncopyrighted to train the decoder.

During the data preprocessing step, the training corpus is tokenized, concatenated into a single stream, and then divided into chunks of each sequence length. 
Additionally, a beginning-of-sentence (BOS) token is appended at the start of each sequence before it is provided to the encoder. 
For the decoder, an end-of-sequence (EOS) token is added to the end of each sequence.

We optimize our model parameters using a language modeling objective with cross-entropy between the input and reconstructed texts as the loss function.
Training is halted using the following early stopping criterion evaluated on a validation set of 128 instances sampled from Pile Uncopyrighted.
The encoder maps each validation instance to Mamba's latent space, and the decoder generates tokens autoregressively until an end-of-sequence (EOS) token is produced or the maximum set threshold of 300 tokens is reached.
Every 1000 steps, we measure the validation ROUGE F1-score.
If it does not improve by at least 0.1 over 5000 training steps, then training is stopped.

\begin{figure}[!t]
    \centering
        \centering
        \includegraphics{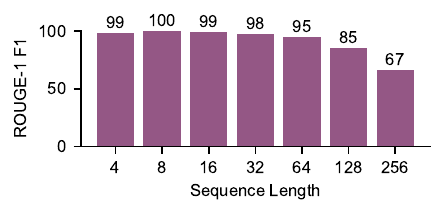}
        \caption{\textbf{Reconstruction Performance Across Sequence Lengths}. ROUGE F1-score for reconstructing text using Mamba (130M). Performance declines sharply as sequence length increases.}
        \label{fig:length}
\end{figure}

\begin{figure}[!t]
        \centering
        \includegraphics[width=\linewidth]{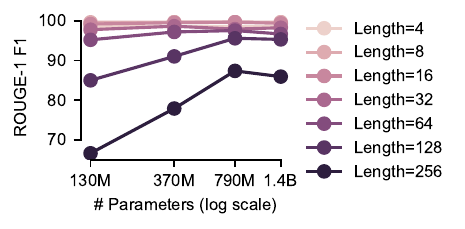}
        \caption{\textbf{Performance vs. Model Size}. ROUGE F1-score as a function of model size, broken down by sequence length. Shorter sequences achieve high performance with fewer parameters, while longer sequences benefit significantly from increased model capacity.}
        \label{fig:scale}
\end{figure}

\begin{figure*}[!th]
    \centering
    \includegraphics{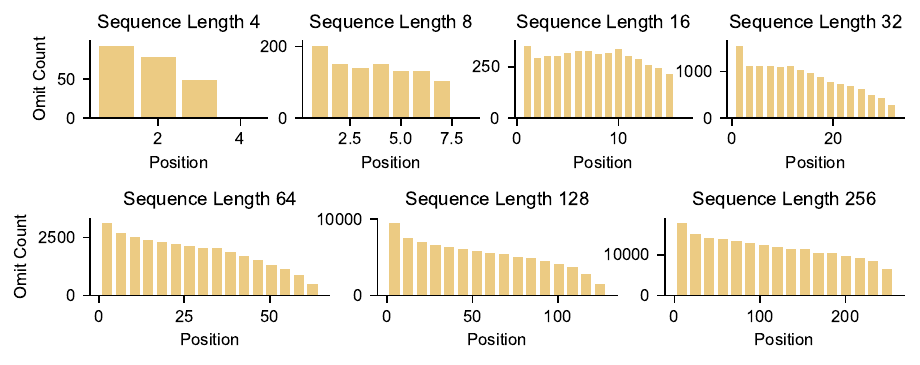}
    \caption{\textbf{Error Positions}. We count the number of reconstruction errors per position in a sequence to assess whether token position impacts reconstruction accuracy.
    We find that later positions have more reconstruction errors than earlier positions.
    130M model errors by position are shown here.
    }
    \label{fig:position}
\end{figure*}

\subsection{Replicating Existing Findings}
\label{sec:validation}
We validate our approach on 700 held-out instances from Pile Uncopyrighted, evaluating whether it replicates existing findings on Mamba's information loss as a function of sequence length, model size, and token position.
To measure information loss across sequence length and model size we use the ROUGE F1-score. 
To measure information loss across token position we use omission rate.

\paragraph{Sequence Length} Like prior work \citep{Jelassi2024RepeatAM, Waleffe2024AnES},  we see a decline in reconstruction fidelity as sequence length increases across model sizes (Figure \ref{fig:scale}).
For instance, for a 130M size model, the reconstruction for the shortest sequence (length 4) is not perfect; it remains very high at 98.6 (Figure \ref{fig:length}). 
However, starting at sequence length 16, the scores begin to drop, with sequences of length 128 and 256 showing notably poor performance, with ROUGE F1-scores of 85.0 and 66.6, respectively.

\paragraph{Model Size} As the number of parameters increases, the performance generally improves, with larger sequence lengths benefiting more from increased model size (Figure \ref{fig:scale}). 
For smaller sequence lengths ($<$16), even the smallest models learn the reconstruction effectively, achieving high performance.
However, for longer sequences (128 and 256), performance improves substantially with an increase in model parameters, showing a much steeper rise in the ROUGE F1-score as the model size grows. 
This suggests that longer sequence lengths require more capacity in the models to adequately encode and decode the sequence information required for reconstruction, reinforcing prior theoretical results \citep{Jelassi2024RepeatAM}.

For sequence lengths 64 and 256, we observe that Mamba 790M outperforms Mamba 1.4B. 
Such non-monotonic behavior has been noted in prior work---for example, \citet{lester-etal-2021-power} report that a T5-Small model can outperform larger T5-Base/Large/XL models under certain prompt-tuning conditions. 
Nonetheless, the overall trend supports the conclusion that larger models are better equipped to handle longer sequences.

\paragraph{Token Position} In line with prior work \citep{wang2025understanding}, we observe a \textit{recency bias} in Mamba's memory, i.e., earlier tokens are omitted at higher rates than recent ones.
We observe this phenomenon across sequence lengths and model sizes.
For example, in Figure \ref{fig:position} we see tokens at the beginning of a sequence being forgotten up to 6 times more than ones at the end of sequences.

\subsection{Token-Level Trends}
\label{sec:token-level}

To understand what types of tokens Mamba tends to retain versus forget in its hidden state, we look at (i) tokens with the top omission rates in natural text, and (ii) omission rates across part-of-speech and named entity categories in annotated data.

\paragraph{Top Omitted Tokens in Natural Text} To investigate what tokens are most frequently omitted in natural text reconstruction, we again use the Pile evaluation dataset across all sequence lengths, and measure ommission rates per token.
To prevent artifacts due to data scarcity, tokens appearing fewer than 100 times in the dataset are excluded. 

Table~\ref{tab:omission_rates} lists the top omitted tokens for a model size of 130M (results for the 1.4B model can be found in  Table~\ref{tab:omission_rates-1.4b}). 
We find that the top 50 omitted tokens include many numbers, letters, and stop words across model sizes.
This provides our first piece of evidence that Mamba has memory issues with math-related texts as numbers and letters correspond to tokens that often take the role of variables.

\paragraph{Part-of-Speech and Named Entities} To investigate how well different parts-of-speech (PoS) and Named Entities (NE) are retained by Mamba, we use the test split of CoNLL-2003.
It contains 3,453 instances of English newswire text, with token-level annotations for PoS and NE tags.
To ensure consistency in sequence length, instances are concatenated up to a fixed length of 256 as it has the highest omission rate.
After excluding categories with less than 100 instances, we compute omission rates across PoS and NE categories, and assess differences between categories using t-tests. 
We apply Bonferroni correction to adjust p-values for multiple comparisons ($\alpha =0.05$).

\begin{table*}[!th]
\small
    \centering
    \resizebox{\linewidth}{!}{%
    \begin{tabular}{lr|lr|lr|lr|lr}
        \toprule
        \textbf{Token} & \textbf{Omit Rate} & 
        \textbf{Token} & \textbf{Omit Rate} & 
        \textbf{Token} & \textbf{Omit Rate} & 
        \textbf{Token} & \textbf{Omit Rate} &
        \textbf{Token} & \textbf{Omit Rate} \\
        \midrule
\textvisiblespace (- & 23.5 & \textvisiblespace times & 14.6 & \textvisiblespace z & 13.5 & \textvisiblespace b & 12.2 & \textvisiblespace less & 11.6 \\
\textvisiblespace 13 & 18.7 & \textvisiblespace s & 14.3 & \textvisiblespace 20 & 13.4 & \textvisiblespace further & 12.1 & 21 & 11.6 \\
\textvisiblespace 11 & 18.3 & \textvisiblespace k & 14.3 & \textvisiblespace 15 & 13.1 & \textvisiblespace Suppose & 12.0 & 62 & 11.5 \\
\textvisiblespace 4 & 15.9 & \textvisiblespace 17 & 14.3 & \textvisiblespace four & 13.0 & \textvisiblespace however & 12.0 & \textvisiblespace 16 & 11.4 \\
Suppose & 15.4 & \textvisiblespace l & 14.2 & \textvisiblespace 12 & 12.8 & \textvisiblespace 8 & 11.9 & )) & 11.4 \\
\textvisiblespace u & 15.3 & \textvisiblespace 7 & 14.1 & 31 & 12.8 & \textvisiblespace 9 & 11.9 & \textvisiblespace either & 11.3 \\
\textvisiblespace w & 15.0 & with & 14.1 & \textvisiblespace 30 & 12.7 & \textvisiblespace d & 11.8 & \textvisiblespace 18 & 11.2 \\
\textvisiblespace o & 14.9 & \textvisiblespace 3 & 13.7 & \textvisiblespace List & 12.6 & \textvisiblespace y & 11.7 & \textvisiblespace similar & 11.0 \\
Category & 14.9 & \textvisiblespace 6 & 13.6 & \textvisiblespace j & 12.6 & 58 & 11.7 & \textvisiblespace above & 10.9 \\
\textvisiblespace 14 & 14.6 & \textvisiblespace 5 & 13.5 & \textvisiblespace n & 12.2 & \textvisiblespace c & 11.7 & and & 10.9 \\
        \bottomrule
    \end{tabular}
    }
    \caption{\textbf{Top 50 Omitted Tokens.} The 50 most forgotten tokens by Mamba (130M) on the Pile across sequence lengths.
    }
    \label{tab:omission_rates}
\end{table*}

\begin{figure}[tb]
        \centering
        \includegraphics{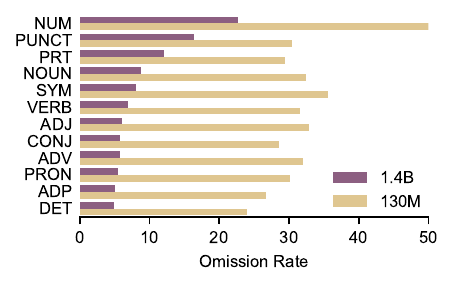} %
        \caption{\textbf{Part-of-Speech.} 
        Omission rates by PoS type on CoNLL-2003.
        Numbers have the highest omission rate across model sizes.
        }
        \label{fig:pos_cat}
\end{figure}

For PoS tags, we find numbers have the highest omission rate across model sizes.
For a 130M model, the omission rate is 50.8\% and for a 1.4B model the omission rate is 22.7\% (Figure~\ref{fig:pos_cat}).
We find a statistically significant difference between numbers and all other PoS categories across model sizes (Tables~\ref{tab:pos_stat_results} and \ref{tab:pos_stat_results-1.4b} in the Appendix).
This reinforces our result from the previous experiment that Mamba's hidden state struggles to retain numbers. 
For the 130M model, we also find statistically significant differences between a few other PoS pairs.
For the 1.4B model we find statistically significant differences in omission rates for punctuations, particles, and nouns (which are the 3 categories that rank just below numbers as the most frequently omitted) and many other categories.

\begin{figure}[tb]
        \centering
        \includegraphics{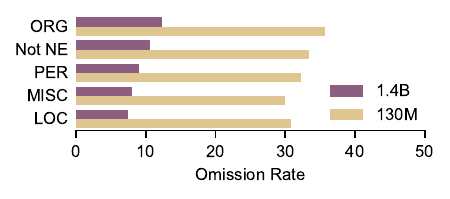} %
        \caption{\textbf{Named Entities.} 
        Omission rates by NE type on CoNLL-2003.
        Organizations have the highest omission rate for both models.
        }
        \label{fig:ner_cat}
\end{figure}

For named entities, we find organizations have the highest omission rate across model sizes.
For the 130M model, the omission rate is 35.8\% while it is 12.3\% for the 1.4B model (Figure \ref{fig:ner_cat}).
We find a statistically significant difference between organizations and other NE categories (Tables \ref{tab:ner_stat_results} and \ref{tab:ne_stat_results-1.4b} in the Appendix).
For the 1.4B model, we also find statistically significant differences between Location, which had the lowest omission rate, and non-named entities.

\subsection{Sequence-Level Trends}
\label{sec:sequence-level}
In order to understand what types of sequences Mamba tends to retain versus forget in its hidden state, we look at reconstruction fidelity across (i) different subsets of Pile Uncopyrighted, (ii) different dialects, and (iii) synthetic datasets.

\paragraph{Pile Subsets} We test how well Mamba's hidden state preserves different types of information by evaluating reconstruction performance across diverse textual domains. 
We sample 700 sequences each from nine subsets of the Pile Uncopyrighted—Common Crawl, ArXiv, NIH, GitHub, PubMed Central, Stack Exchange, Enron Email, Free Law, and DM Mathematics—capturing a broad range of linguistic characteristics, including web text, scientific writing, code, legal documents, and mathematical content.

\begin{figure*}[!th]
    \centering
    \includegraphics{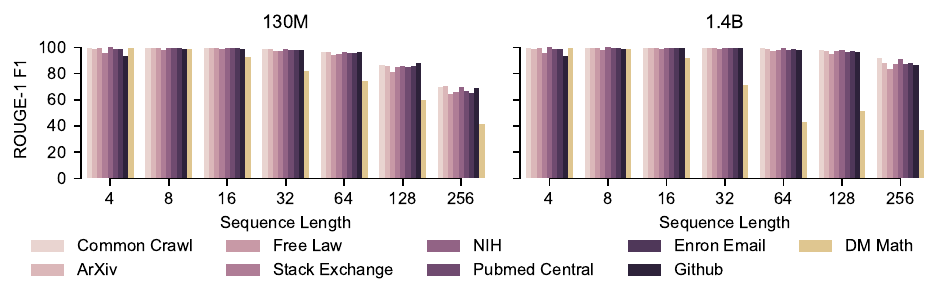}
    \caption{\textbf{Reconstruction Errors by Input Source.} Reconstruction fidelity across subsets of Pile for Mamba.
    As sequence length increases, reconstruction for the DM Math subset deteriorates the most compared to other subsets.
    }
    \label{fig:dist}
\end{figure*}

Across model sizes, we find that the fidelity of reconstruction deteriorates the most for mathematical data as sequence length increases.
For the 130M model, the ROUGE F1-score for the DM Mathematics subset is 99.9 for a sequence length of 4, which is on par with the other subsets.
At a sequence length of 256, reconstruction performance on DM Mathematics drops sharply, with an F1-score of 41.6---substantially lower than other subsets, such as Common Crawl, which achieves 69.7 (Figure \ref{fig:dist}).   
In contrast, the remaining subsets show relatively similar performance to one another.
This further supports our finding from token-level experiments that Mamba's hidden state struggles to retain information from mathematical inputs.

\paragraph{Dialects} To evaluate if dialect impacts Mamba's retention, we use the AAVE/SAE paired dataset \citep{groenwold-etal-2020-investigating}, which contains parallel instances in African American Vernacular English (AAVE) and Standard American English (SAE). 
For consistency, instances are concatenated to a fixed maximum length, ranging from 4--256 tokens.

\begin{figure}
        \centering
        \includegraphics{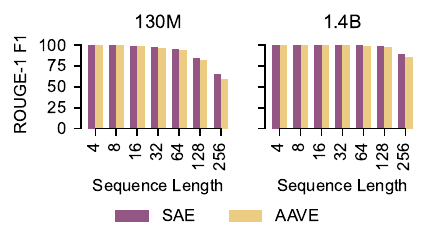}
        \caption{\textbf{Dialect.} Compared to Standard American English (SAE), the reconstruction fidelity of African American Vernacular (AAVE) degrades more with increased sequence length.
        }
        \label{fig:race}
\end{figure}

For short sequences, ROUGE F1-scores for reconstruction are high and nearly identical for SAE and AAVE. 
For a 130M model at a sequence length of 4, SAE achieves 99.7 while AAVE is at 99.6 ($\Delta = 0.1$; Figure \ref{fig:race}). 
However, as sequence length increases, the performance gap between the two widens. 
At a sequence length of 256 tokens, SAE reconstruction F1 declines to 65.3, while AAVE falls further to 59.7 (\(\Delta = 5.6\)). 
We see similar trends across model sizes (Figure~\ref{fig:race}).

\paragraph{Synthetic Numeric Sequences} We assess how well purely numerical sequences can be reconstructed by creating a synthetic evaluation dataset. 
For each sequence length (4–256 tokens), we randomly sample numeric tokens from Mamba’s vocabulary to generate 1K instances of the corresponding length. 

\begin{figure}[!th]
    \centering
    \includegraphics{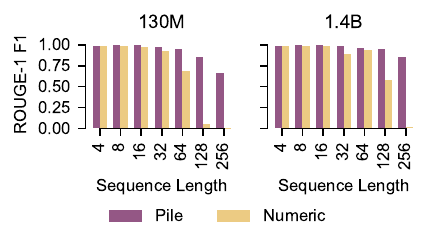}
    \caption{\textbf{Numerical Sequences.} Comparison of reconstruction fidelity of randomly sampled numeric sequences versus natural sequences from the Pile.}
    \label{fig:numbers}
\end{figure}

We found that accuracy declines significantly more with length compared to natural text. 
For a 130M-parameter model at 256 tokens, the F1-score drops from 66.5 for standard text to just 0.5 for numbers (Figure \ref{fig:numbers}).
We see similar trends across model sizes.

\paragraph{Difference Between Reference and Generated Numbers}
To understand the character-level deviations involved in retaining numeric information, we compute the Levenshtein distance between incorrectly generated numeric tokens and their corresponding references on the CoNLL-2003 and Pile evaluation sets.
Across datasets and model sizes, the median reference token length ranged from 2--3, with a median edit distance of 2 (Appendix \ref{app:num}).
These values show that for errors on numerical tokens, most characters do not match.

To understand the magnitude of this difference between reference and generated numbers, we extract numbers from incorrect instances and compute the Mean Absolute Percentage Error (MAPE), which expresses the mean absolute error as a ratio of the reference values, providing a scale-independent metric for error.
On the CoNLL-2003 test set, the generated numeric token magnitudes for the 130M model deviate from the reference tokens by an average of 8.4$\times$ their magnitude. 
The MAPE decreases steadily with increasing model size, reaching 1.9 at 1B parameters. 
When evaluated on the Pile dataset, the MAPE increases with chunk length across model sizes, rising from approximately 0 at a chunk size of 4 to 38.2 at a chunk size of 256 (Appendix \ref{app:num}).
These results suggest that larger models better preserve the magnitudes of numeric tokens, whereas numeric tokens in longer sequences are retained poorly with high deviation.

\paragraph{Repeated Tokens} We test whether Mamba’s hidden state retains both token identity and repetition count by creating a synthetic dataset of 2K randomly sampled tokens from its vocabulary.
Each token is repeated $n$ times, where $n$ ranges from 4 to 256. 
For a 130M model, repeated tokens were correctly generated $>90\%$ of times across chunk sizes (Table \ref{tab:repeat}). 
However, only at a sequence length of 4 did the model consistently reproduce the correct number of repetitions. 
For larger sequence lengths, the model instead continuing generation until reaching the maximum token limit of 300.

\begin{figure}[h]
    \centering
        \centering
        \includegraphics{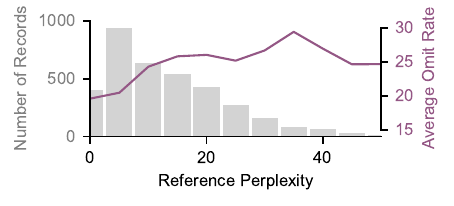}
        \caption{\textbf{Perplexity vs. Omission Rates.} We observe that as the perplexity of input sequences increases, so does the omission rate of their reconstructions.
        } 
        \label{fig:perplexity}
\end{figure}

\label{sec:mem_perp}
\paragraph{Memory and Perplexity} We then examine whether Mamba's hidden state has more difficulty retaining information about sequences that are less likely under the model, i.e., higher perplexity, by correlating reference perplexity with omission rates in the evaluation subset of the Pile.
For a 130M Mamba model at a sequence length of 256, we observe that as the perplexity of input sequences increases, so does the omission rates of their reconstructions (Figure \ref{fig:perplexity}).
Thus, it is harder for Mamba to accurately store information about higher perplexity sequences.

\subsection{Training Corpus Analysis}
To investigate why certain types of tokens and sequences are more prone to omission in Mamba, we examine its training data. 
We randomly sample 200K instances from Pile Uncopyrighted by first randomly selecting a partition of the dataset, and then applying reservoir sampling. 
We annotate this sample using the Stanza part-of-speech tagger.

We find that numerical tokens (NUM) exhibit the lowest count per unique token among all categories evaluated, 
while, determiners (DET)---which have the lowest omission rates---have the highest count per unique token (Table~\ref{tab:pos_stats}).
This shows an association between omission rates and pre-training frequencies.
\begin{table}[ht]
\small
\centering
\begin{tabular}{lccc}
\toprule
\textbf{PoS} & \textbf{Total} & \textbf{Unique} & \textbf{Ratio} \\
\midrule
DET & 14M & 887 & 16K \\
PRON & 7.6M & 2.4K & 3.1K \\
ADP & 18M & 5.9K & 3.0K \\
PUNCT & 43M & 73K & 584 \\
ADV & 5.6M & 17K & 325 \\
SYM & 6.0M & 21K & 293 \\
VERB & 16M & 79K & 205 \\
ADJ & 12M & 93K & 130 \\
NOUN & 47M & 158K & 30 \\
NUM & 9.0M & 643K & 14 \\
\bottomrule
\end{tabular}
\caption{\textbf{Part-of-Speech Distribution in Pile Sample.} In a 200K random sample of the Pile Uncopyrighted, numbers (NUM) have the lowest count per unique token, which may contribute to their poor retention. %
}
\label{tab:pos_stats}
\end{table}

\section{Discussion}
\label{sec:discussion}

Our main contribution is an evaluation of the selective memory of the pretrained Mamba \emph{language models}.
We emphasize this term to highlight our focus on the impact language model pretraining has on information retention, as opposed to fundamental limitations of the Mamba architecture.
Our findings show that Mamba LMs are especially prone to forgetting certain types of information (e.g., numeric tokens) and this appears to be driven in part by occurrence in pretraining data.
In addition to our analysis, our work contributes a general framework for diagnosing memory limitations in LMs with fixed-size hidden states, offering a tool for future research. 
As LMs are often used as black-box systems, understanding their limitations is a critical first step toward mitigating them. 
This paper aligns with existing evaluation-focused research in NLP %
\citep{hessel-etal-2023-androids, lin-etal-2022-truthfulqa, moghe-etal-2023-extrinsic, selvam-etal-2023-tail} that sheds light on issues and guides future work.

Building on our findings, we suggest that the conventional pretraining pipeline may not be ideally suited to the inductive biases of SSM-based language models.
One aspect to reconsider is tokenization. 
Recent work has demonstrated that byte-level SSM language modeling can outperform transformers on various benchmarks \citep{Wang2024MambaByteTS}, indicating that tokenization-free approaches may provide a more natural inductive bias for SSMs \citep{gu2025tradeoffs}. 
Tokenization choice also impacts numerical reasoning; \citet{Singh2024TokenizationCT} demonstrate that tokenizing numbers as single digits yields better performance on mathematical tasks. %
Similarly, \citet{Yang2024NumberCN} show that right-to-left tokenization can enhance numeracy.

The pretraining objective itself is yet another direction to revisit. 
Auxiliary reconstruction losses and memory-augmenting self-supervised objectives improve memory retention in RNN and LSTM models \citep{Zhang2021LearningTR, Trinh2018LearningLD}. 
Similarly, arithmetic-aware pretraining strategies, such as contrastive learning and task-specific auxiliary losses, enhance numerical reasoning in LLMs~\citep{petrak-etal-2023-arithmetic, li-etal-2022-seeking}. 
We hope that future work will address the selective memory issues in SSM language models by exploring pre-training strategies tailored to SSM architectures.

\section{Conclusion}

In this work, we characterized the types of tokens (e.g., parts of speech, named entities) and sequences (e.g., code, mathematical content) that are most susceptible to being forgotten by Mamba LMs.
By training auto-encoders to reconstruct inputs directly from Mamba’s hidden states, we were able to assess information retention without relying on explicit memorization cues like previous works.
Our results showed substantially greater information loss for math-related tokens as well as for organization names and non-standard English dialects. 
These findings highlight that Mamba's memory is clearly selective, which could pose a problem for applying it to tasks that require exactly recalling the types of tokens it has a tendency to forget, e.g., math reasoning, information retrieval, and open-domain dialogue tasks.
By identifying these weaknesses, our work provides motivation and direction for future research into methods that control Mamba's selective memory so that it is able to better retain pertinent information.
We release our full implementation at \url{https://github.com/dataminr-ai/ouroboros} to support future work.

\section{Limitations}
\paragraph{Decoder Errors} A limitation of our methodology is the potential for confounding between decoder errors and information loss in Mamba's hidden states. 
Since we train a decoder for text reconstruction,  it is possible some of the reconstruction omissions we find are due to decoder errors as well as information lacking in Mamba’s hidden states. 
Nevertheless, the successful replication of prior work with our method provides strong validation of our approach, suggesting that any impact from decoder errors is not significant enough to invalidate our core findings.

\paragraph {Single SSM Language Model Family} We only evaluate on one SSM language model family, Mamba. 
The reason we chose these models is that, similar to how ChatGPT has been studied as a representative of state-of-the-art transformer LLMs \citep{gao-etal-2023-chatgpt, laskar-etal-2023-systematic, bang-etal-2023-multitask, jang-lukasiewicz-2023-consistency}, the pretrained Mamba models are representatives of state-of-the-art SSM LLMs. 
Additionally, there are no other reasonable pure SSM language models available for comparison.

\paragraph{Single Linear Architecture} Our study focuses solely on one linear recurrent architecture, SSMs, though our approach can be applied to other architectures as well, such as RWKV \citep{peng-etal-2023-rwkv}.
While exploring information retention across these different architectures is an important direction for future research (and if our paper is accepted, we will make our code publicly available to facilitate such work), this paper already required a substantial amount of resources to produce. For instance training the 1.4B decoder required multiple weeks on a 48GB A6000 GPU.

\paragraph{Solutions}  We do not attempt to solve the selective memory issues we find in Mamba, though we discuss some possible options in Section ~\ref{sec:discussion}. 
However, identifying a problem is an essential first step toward developing an effective solution. 
Developing solutions, particularly for complex behaviors in large language models, is non-trivial and typically requires targeted, follow-up research. 
Our paper falls within the well-established tradition in NLP of evaluation-focused work \citep{hessel-etal-2023-androids, lin-etal-2022-truthfulqa, moghe-etal-2023-extrinsic, selvam-etal-2023-tail}, and aims to shed light on an important issue that can guide future efforts toward mitigate identified issues.

\paragraph{Downstream Task} We do not evaluate on downstream tasks. 
However, we identify a fundamental issue with Mamba: its difficulty in retaining crucial information, such as numbers, from context.
This is an essential skill for techniques that are now standard in LLM reasoning, such as task decomposition \citep{Wei2022ChainOT, Yao2023TreeOT}. 
While downstream evaluation is beyond the scope of this paper, we encourage future researchers to explore this area.

\paragraph{Dataset} We use samples from Pile Uncopyrighted to train our autoencoder and estimate the part-of-speech composition of Mamba’s training data. 
However, there might be some distribution shift between Pile Uncopyrighted and The Pile, the original training dataset for the Mamba language models.
However, this is unavoidable as the full Pile is no longer available for use.

\paragraph{Training Corpus Analysis} While we demonstrate a strong connection between the forgetting of numerical data and the occurrence of numbers in Mamba's training corpus, we are unable to establish a similar connection for all types of information loss, such as non-standard dialects. 
This is due to the lack of available taggers for dialects, in contrast to the well-established taggers for parts-of-speech.

\bibliography{custom}

\clearpage
\appendix
\setcounter{table}{0}
\renewcommand{\thetable}{A\arabic{table}}
\setcounter{figure}{0}
\renewcommand{\thefigure}{A\arabic{figure}}

\newpage

\section{Mamba Details}
\begin{table}[!th]
\centering
\small
\begin{tabular}{@{}cccc@{}}
\toprule
\textbf{Model} & \textbf{Layers} & \textbf{$\beta_S$} & \textbf{$\beta_c$} \\ \midrule
130M           & 24              & {[}1536, 16{]}     & {[}1536,4{]}                                                            \\
370M           & 48              & {[}2048,16{]}      & {[}2048,4{]}                                                            \\
790M           & 48              & {[}3072,16{]}      & {[}3072,4{]}                                                            \\
1.4B           & 48              & {[}4096,16{]}      & {[}4096,4{]}                                                            \\ \bottomrule
\end{tabular}
\caption{\textbf{Hidden State Sizes.} The encoder maps input texts into a hidden state with two parts: (i) an SSM State, $\beta_S$, and (ii) a Convolutional State, $\beta_C$.
The sizes $\beta_S$ and $\beta_C$ per layer for encoders of various sizes between 130m and 1.4B are shown.
}

\label{table:size}
\end{table}
\section{Additional 130M Results}
\begin{table}[H]
    \centering
    \small
    \resizebox{\columnwidth}{!}{%
    \begin{tabular}{lccc}
        \toprule
        \bf{Pair} & \bf{T-Statistic} & \bf{Adjusted P-Value} \\
        \midrule
NUM - SYM   & -4.4  & 8.8e-04 \\
NUM - ADJ   & -15.4 & 5.1e-51 \\
NUM - NOUN  & -25.1 & 1.8e-135 \\
NUM - ADV   & -11.3 & 1.9e-27 \\
NUM - VERB  & -20.8 & 4.6e-92 \\
NUM - PUNCT & -22.8 & 1.6e-110 \\
NUM - PRON  & -12.3 & 1.1e-32 \\
NUM - PRT   & -8.8  & 7.8e-17 \\
NUM - CONJ  & -11.7 & 1.3e-29 \\
NUM - ADP   & -26.1 & 3.0e-144 \\
NUM - DET   & -24.8 & 1.1e-129 \\
SYM - DET   & -3.8  & 1.1e-02 \\
ADJ - ADP   & -5.6  & 1.9e-06 \\
ADJ - DET   & -7.3  & 2.1e-11 \\
NOUN - ADP  & -7.6  & 1.7e-12 \\
NOUN - DET  & -9.1  & 4.6e-18 \\
ADV - ADP   & -3.5  & 2.7e-02 \\
ADV - DET   & -5.1  & 1.9e-05 \\
VERB - ADP  & -5.3  & 6.6e-06 \\
VERB - DET  & -7.2  & 3.4e-11 \\
PUNCT - ADP & -4.2  & 1.9e-03 \\
PUNCT - DET & -6.3  & 2.4e-08 \\
PRON - DET  & -3.9  & 5.6e-03 \\
        \bottomrule
    \end{tabular}
    }
    \caption{
    \textbf{Statistical Tests (Part-of-Speech)}.
    Part-of-speech categories that have statistically different omission rates based on pair-wise t-tests at $\alpha=0.05$ with Bonferroni corrected p-values for Mamba (130M).}
    \label{tab:pos_stat_results}
\end{table}

\begin{table}[H]
    \centering
    \small
    \resizebox{\columnwidth}{!}{%
    \begin{tabular}{lccc}
        \toprule
        \bf{Pair} & \bf{T-Statistic} &  \bf{Adjusted P-Value} \\
        \midrule
 Organization - Location & -3.4 & 6.3e-03 \\
Organization - Misc     & -3.2 & 1.5e-02 \\
        \bottomrule
    \end{tabular}
    }
    \caption{\textbf{Statistical Tests (Named Entity Recognition)}. Named entity categories that have statistically different commission rates based on pair-wise t-tests at $\alpha=0.05$ with Bonferroni corrected p-values for Mamba (130M).}
    \label{tab:ner_stat_results}
\end{table}

\begin{table}[!th]
\small
    \centering
    \begin{tabular}{c c c}
        \toprule
        \textbf{Length} & \textbf{Present (\%)} & \textbf{Repeat Mode} \\
        \midrule
        4   & 99.9  & 4   \\
        8   & 99.8  & 300 \\
        16  & 100.0 & 300 \\
        32  & 100.0 & 300 \\
        64  & 99.9  & 300 \\
        128 & 99.7  & 300 \\
        256 & 92.1  & 300 \\
        \bottomrule
    \end{tabular}
    \caption{\textbf{Repeated Tokens.} The percentage of generations that contained the correct repeated tokens and the modal repetition count observed in those generations.}
    \label{tab:repeat}
\end{table}

\section{Additional 1.4B Results}

\begin{table*}[!t]
\small
    \centering
    \resizebox{\linewidth}{!}{%
    \begin{tabular}{lr|lr|lr|lr|lr}
        \toprule
        \textbf{Token} & \textbf{Omit Rate} & 
        \textbf{Token} & \textbf{Omit Rate} & 
        \textbf{Token} & \textbf{Omit Rate} & 
        \textbf{Token} & \textbf{Omit Rate} &
        \textbf{Token} & \textbf{Omit Rate} \\
        \midrule
\textvisiblespace (- & 43.8 & \textvisiblespace w & 15.1 & \textvisiblespace 7 & 13.2 & \textvisiblespace four & 10.7 & w & 9.6 \\
What & 23.5 & \textvisiblespace Suppose & 14.8 & \textvisiblespace l & 13.0 & \textvisiblespace d & 10.6 & \textvisiblespace t & 9.5 \\
Suppose & 23.0 & \textvisiblespace z & 14.6 & \textvisiblespace 5 & 12.7 & \textvisiblespace 20 & 10.4 & \textvisiblespace prime & 9.5 \\
\textvisiblespace List & 21.9 & \textvisiblespace 17 & 14.4 & \textvisiblespace b & 12.4 & \textvisiblespace n & 10.3 & \textvisiblespace Let & 9.4 \\
\textvisiblespace What & 17.8 & \textvisiblespace 6 & 13.9 & 21 & 11.4 & \textvisiblespace 15 & 10.3 & \textvisiblespace f & 9.4 \\
\textvisiblespace 13 & 17.0 & \textvisiblespace u & 13.7 & \textvisiblespace 9 & 11.4 & )) & 10.0 & \textvisiblespace g & 9.1 \\
\textvisiblespace 11 & 17.0 & \textvisiblespace 14 & 13.6 & \textvisiblespace 12 & 11.2 & \textvisiblespace 16 & 9.9 & \textvisiblespace h & 9.1 \\
\textvisiblespace 4 & 16.8 & \textvisiblespace j & 13.3 & \textvisiblespace 3 & 11.2 & \textvisiblespace 2 & 9.9 & ? & 8.7 \\
\textvisiblespace times & 15.6 & \textvisiblespace y & 13.3 & \textvisiblespace c & 11.0 & \textvisiblespace 18 & 9.7 & ** & 8.7 \\
\textvisiblespace o & 15.1 & \textvisiblespace s & 13.2 & \textvisiblespace k & 10.8 & \textvisiblespace 8 & 9.6 & \textvisiblespace 1 & 8.4 \\
        \bottomrule
    \end{tabular}
    }
    \caption{\textbf{Top 50 Omitted Tokens.} The top 50 tokens most forgotten words in
natural text across sequence lengths for Mamba (1.4B).}
    \label{tab:omission_rates-1.4b}
\end{table*}

\begin{table}[H]
    \centering
    \resizebox{\columnwidth}{!}{%
    \begin{tabular}{lcc}
        \toprule
        \textbf{Pair} & \textbf{T-Statistic} & \textbf{Adjusted P-Value} \\
        \midrule
NUM - PUNCT  & -8.6  & 6.5e-16 \\
NUM - PRT    & -5.2  & 1.2e-05 \\
NUM - NOUN   & -28.0 & 1.0e-167 \\
NUM - SYM    & -5.0  & 3.4e-05 \\
NUM - VERB   & -23.3 & 5.4e-115 \\
NUM - ADJ    & -18.7 & 2.4e-74 \\
NUM - CONJ   & -11.0 & 7.4e-26 \\
NUM - ADV    & -12.7 & 5.4e-35 \\
NUM - PRON   & -12.8 & 2.4e-35 \\
NUM - ADP    & -26.5 & 1.2e-148 \\
NUM - DET    & -21.5 & 8.6e-98 \\
PUNCT - NOUN & -15.9 & 7.7e-55 \\
PUNCT - VERB & -15.1 & 2.4e-49 \\
PUNCT - ADJ  & -12.8 & 2.5e-35 \\
PUNCT - CONJ & -7.7  & 9.5e-13 \\
PUNCT - ADV  & -9.0  & 2.6e-17 \\
PUNCT - PRON & -9.1  & 8.1e-18 \\
PUNCT - ADP  & -18.7 & 2.4e-75 \\
PUNCT - DET  & -15.3 & 1.6e-50 \\
PRT - VERB   & -4.0  & 3.6e-03 \\
PRT - ADJ    & -4.7  & 2.2e-04 \\
PRT - CONJ   & -4.0  & 5.1e-03 \\
PRT - ADV    & -4.3  & 1.1e-03 \\
PRT - PRON   & -4.5  & 4.6e-04 \\
PRT - ADP    & -6.4  & 1.4e-08 \\
PRT - DET    & -6.1  & 9.0e-08 \\
NOUN - VERB  & -4.0  & 4.1e-03 \\
NOUN - ADJ   & -4.5  & 5.1e-04 \\
NOUN - ADV   & -3.4  & 4.7e-02 \\
NOUN - PRON  & -3.6  & 1.8e-02 \\
NOUN - ADP   & -8.5  & 1.1e-15 \\
NOUN - DET   & -6.9  & 3.4e-10 \\
VERB - ADP   & -4.2  & 2.2e-03 \\
VERB - DET   & -3.6  & 2.1e-02 \\
        \bottomrule
    \end{tabular}
    }
    \caption{
    \textbf{Statistical Tests (Part-of-Speech)}.
    Part-of-speech categories that have statistically different omission rates based on pair-wise t-tests at $\alpha=0.05$ with Bonferroni corrected p-values for Mamba (1.4B).}
    \label{tab:pos_stat_results-1.4b}
\end{table}

\begin{table}[H]
    \centering
    \resizebox{\columnwidth}{!}{%
    \begin{tabular}{lcc}
        \toprule
        \textbf{Pair} & \textbf{T-Statistic} & \textbf{Adjusted P-Value} \\
        \midrule
Organization - Person   & -3.9 & 1.0e-03 \\
Organization - Misc     & -3.6 & 3.7e-03 \\
Organization - Location & -5.2 & 1.8e-06 \\
Not NE - Location       & -4.4 & 9.8e-05 \\
        \bottomrule
    \end{tabular}
    }
    \caption{
    \textbf{Statistical Tests (Named Entities)}.
    Named entity categories with statistically different omission rates based on pair-wise t-tests at $\alpha=0.05$ with Bonferroni corrected p-values for Mamba (1.4B).}
    \label{tab:ne_stat_results-1.4b}
\end{table}

\clearpage
\section{How Different are Reference and Generated Numbers?}
\label{app:num}

\paragraph{Character difference} The lengths of reference numeric tokens that are incorrectly reconstructed are shown in Table \ref{tab:cr_length_distribution}, and the Levenshtein distance between reference and generated tokens are showns in Table \ref{tab:levenshtein_edit_distance}.

\paragraph{Magnitude difference} The Mean Absolute Percentage Error (MAPE) for numbers in incorrectly generated numeric tokens across CoNNL-2003 and Pile evaluation datasets are shown in Table \ref{tab:numeric_distance}.

\begin{table}[!h]
\centering
\small
\begin{tabular}{llccc}
\toprule
Dataset & Model & Chunk & Mean & Median \\
\midrule
connl & 130m & 256 & 1.9 & 2.0 \\
connl & 370m & 256 & 1.9 & 2.0 \\
connl & 790m & 256 & 1.8 & 2.0 \\
connl & 1b & 256 & 1.9 & 2.0 \\
\midrule
pile & 130m & 4 & 4.1 & 2.0 \\
pile & 130m & 8 & 1.7 & 2.0 \\
pile & 130m & 16 & 1.8 & 2.0 \\
pile & 130m & 32 & 2.0 & 2.0 \\
pile & 130m & 64 & 2.0 & 2.0 \\
pile & 130m & 128 & 2.2 & 2.0 \\
pile & 130m & 256 & 2.2 & 2.0 \\
\midrule
pile & 370m & 4 & 0.0 & 0.0 \\
pile & 370m & 8 & 1.6 & 2.0 \\
pile & 370m & 16 & 2.0 & 2.0 \\
pile & 370m & 32 & 2.2 & 2.0 \\
pile & 370m & 64 & 1.9 & 2.0 \\
pile & 370m & 128 & 2.1 & 2.0 \\
pile & 370m & 256 & 2.2 & 2.0 \\
\midrule
pile & 790m & 4 & 2.0 & 2.0 \\
pile & 790m & 8 & 1.7 & 2.0 \\
pile & 790m & 16 & 1.8 & 2.0 \\
pile & 790m & 32 & 1.9 & 2.0 \\
pile & 790m & 64 & 1.9 & 2.0 \\
pile & 790m & 128 & 2.0 & 2.0 \\
pile & 790m & 256 & 2.2 & 2.0 \\
\midrule
pile & 1b & 4 & 1.8 & 2.0 \\
pile & 1b & 8 & 1.8 & 2.0 \\
pile & 1b & 16 & 2.0 & 2.0 \\
pile & 1b & 32 & 2.1 & 2.0 \\
pile & 1b & 64 & 1.9 & 2.0 \\
pile & 1b & 128 & 2.0 & 2.0 \\
pile & 1b & 256 & 2.2 & 2.0 \\
\midrule
\bottomrule
\end{tabular}
\caption{\textbf{Edit Distance.} Levenshtein distance between mismatching reference and generated numeric tokens on the CoNNL-2003 and Pile evaluation datasets }
\label{tab:levenshtein_edit_distance}
\end{table}

\begin{table}[h]
\centering
\small
\begin{tabular}{llccc}
\toprule
Dataset & Model & Chunk & Mean & Median  \\
\midrule
connl & 130m & 256 & 2.5 & 3.0 \\
connl & 370m & 256 & 2.6 & 3.0 \\
connl & 790m & 256 & 2.5 & 3.0 \\
connl & 1b & 256 & 2.5 & 2.0 \\
\midrule
pile & 130m & 4 & 2.1 & 2.0 \\
pile & 130m & 8 & 2.0 & 2.0 \\
pile & 130m & 16 & 1.9 & 2.0 \\
pile & 130m & 32 & 2.1 & 2.0 \\
pile & 130m & 64 & 2.1 & 2.0 \\
pile & 130m & 128 & 2.2 & 2.0 \\
pile & 130m & 256 & 2.1 & 2.0 \\
\midrule
pile & 370m & 4 & 0.0 & 0.0 \\
pile & 370m & 8 & 2.1 & 2.0 \\
pile & 370m & 16 & 2.2 & 2.0 \\
pile & 370m & 32 & 2.2 & 2.0 \\
pile & 370m & 64 & 1.9 & 2.0 \\
pile & 370m & 128 & 2.0 & 2.0 \\
pile & 370m & 256 & 2.1 & 2.0 \\
\midrule
pile & 790m & 4 & 3.0 & 3.0 \\
pile & 790m & 8 & 2.1 & 2.0 \\
pile & 790m & 16 & 2.1 & 2.0 \\
pile & 790m & 32 & 2.0 & 2.0 \\
pile & 790m & 64 & 1.9 & 2.0 \\
pile & 790m & 128 & 1.9 & 2.0 \\
pile & 790m & 256 & 2.1 & 2.0 \\
\midrule
pile & 1b & 4 & 2.4 & 2.5 \\
pile & 1b & 8 & 2.1 & 2.0 \\
pile & 1b & 16 & 2.0 & 2.0 \\
pile & 1b & 32 & 2.2 & 2.0 \\
pile & 1b & 64 & 1.8 & 2.0 \\
pile & 1b & 128 & 2.0 & 2.0 \\
pile & 1b & 256 & 2.0 & 2.0 \\
\bottomrule
\end{tabular}
\caption{\textbf{Length of Error Instance.} Length of incorrectly reconstructed reference tokens in CoNNL-2003 and Pile evaluation datasets}
\label{tab:cr_length_distribution}
\end{table}

\clearpage
\begin{table}[H]
\centering
\small
\begin{tabular}{llcc}
\toprule
\textbf{Dataset} & \textbf{Model} & \textbf{Chunk} &\textbf{ MAPE} \\
\midrule
connl & 130m & 256 & 8.4181 \\
connl & 370m & 256 & 5.9192 \\
connl & 790m & 256 & 2.7187 \\
connl & 1b & 256 & 1.9930 \\
connl & \textit{Avg} & 256 & \textbf{4.7622} \\
\midrule
pile & 130m & 4 & 0.0005 \\
pile & 370m & 4 & 0.0000 \\
pile & 790m & 4 & 0.0000 \\
pile & 1b & 4 & 0.0002 \\
pile & \textit{Avg} & 4 & \textbf{0.0002} \\
\midrule
pile & 130m & 8 & 0.0013 \\
pile & 370m & 8 & 0.0032 \\
pile & 790m & 8 & 0.0026 \\
pile & 1b & 8 & 0.0036 \\
pile & \textit{Avg} & 8 & \textbf{0.0027} \\
\midrule
pile & 130m & 16 & 0.0439 \\
pile & 370m & 16 & 0.0455 \\
pile & 790m & 16 & 0.0214 \\
pile & 1b & 16 & 0.0620 \\
pile & \textit{Avg} & 16 & \textbf{0.0432} \\
\midrule
pile & 130m & 32 & 0.2199 \\
pile & 370m & 32 & 0.1697 \\
pile & 790m & 32 & 0.2208 \\
pile & 1b & 32 & 0.1671 \\
pile & \textit{Avg} & 32 & \textbf{0.1944} \\
\midrule
pile & 130m & 64 & 0.6067 \\
pile & 370m & 64 & 0.2211 \\
pile & 790m & 64 & 0.2175 \\
pile & 1b & 64 & 0.4596 \\
pile & \textit{Avg} & 64 & \textbf{0.3762} \\
\midrule
pile & 130m & 128 & 2.5763 \\
pile & 370m & 128 & 1.0526 \\
pile & 790m & 128 & 0.5142 \\
pile & 1b & 128 & 0.4900 \\
pile & \textit{Avg} & 128 & \textbf{1.1583} \\
\midrule
pile & 130m & 256 & 89.7517 \\
pile & 370m & 256 & 5.9819 \\
pile & 790m & 256 & 3.2479 \\
pile & 1b & 256 & 53.8413 \\
pile & \textit{Avg} & 256 & \textbf{38.2057} \\
\bottomrule
\end{tabular}
\caption{\textbf{Magnitude Differences.} Mean Absolute Percentage Error (MAPE) for numbers in incorrectly generated numeric tokens across CoNNL-2003 and Pile evaluation datasets.
}
\label{tab:numeric_distance}
\end{table}

\end{document}